\pdfoutput=1

\documentclass[11pt]{article}

\usepackage[]{EMNLP2023}

\usepackage{times}
\usepackage{latexsym}

\usepackage[T1]{fontenc}

\usepackage[utf8]{inputenc}

\usepackage{microtype}

\usepackage{inconsolata}

\usepackage{graphicx}
\usepackage{amsmath}
\usepackage{physics}
\usepackage{amsmath}
\usepackage{tikz}
\usepackage{mathdots}
\usepackage{yhmath}
\usepackage{cancel}
\usepackage{color}
\usepackage{siunitx}
\usepackage{array}
\usepackage{multirow}
\usepackage{amssymb}
\usepackage{gensymb}
\usepackage{tabularx}
\usepackage{extarrows}
\usepackage{booktabs}
\usetikzlibrary{fadings}
\usetikzlibrary{patterns}
\usetikzlibrary{shadows.blur}
\usetikzlibrary{shapes}
\usepackage{xspace} 
\usepackage{listings}
\usepackage{listingsutf8}
\usepackage{changepage}

\graphicspath{ {./images/} }

\usepackage{todonotes}

\newcommand{\sysname}{AMREx\xspace}

\title{AMREx: AMR for Explainable Fact Verification}

\author{
\textbf{Chathuri Jayaweera},
\textbf{Sangpil Youm},
\textbf{Bonnie Dorr}
\\
University of Florida, Gainesville, FL,USA \\
\texttt{\{chathuri.jayawee, youms, bonniejdorr\}@ufl.edu}}

\begin{document}
\maketitle
\begin{abstract}
With the advent of social media networks and the vast amount of information circulating through them, automatic fact verification is an essential component to prevent the spread of misinformation. It is even more useful to have fact verification systems that provide explanations along with their classifications to ensure accurate predictions. To address both of these requirements, we implement \sysname, an Abstract Meaning Representation (AMR)-based veracity prediction and explanation system for fact verification using a combination of Smatch, an AMR evaluation metric to measure meaning containment and textual similarity, and demonstrate its effectiveness in producing partially explainable justifications using two community standard fact verification datasets, FEVER and AVeriTeC. \sysname surpasses the AVeriTec baseline accuracy showing the effectiveness of our approach for real-world claim verification. It follows an interpretable pipeline and returns an explainable AMR node mapping to clarify the system's veracity predictions when applicable.
We further demonstrate that \sysname output can be used to prompt LLMs to generate natural-language explanations using the AMR mappings as a guide to lessen the probability of hallucinations.
\end{abstract}

\section{Introduction}
With the vast 
amount of information 
circulating on social media and 
the constantly changing
Claims about various topics,
automatic fact verification has become 
crucial for preventing the spread of
misinformation. To 
address this need,
automatic fact-checking task \citep{vlachos-riedel-2014-fact} and several shared tasks have been introduced to encourage NLP researchers to 
develop systems 
that gather Evidence (Fact extraction) 
for 
a given Claim and 
classify it (Fact verification) 
as to its predicted veracity.  Examples include FEVER \citep{thorne-etal-2018-fact, thorne-etal-2019-fever2} and the current AVeriTec task \cite{schlichtkrull2023averitec,schlichtkrull2024averitec}, 
which employ
the labels \verb|Supports|, \verb|Refutes|, \verb|NotEnoughInfo| (NEI) or \verb|ConflictingEvidence/CherryPicking|.

Natural Language Inference (NLI) systems, which assess whether
a premise 
semantically entails
a given hypothesis \citep{bowman-etal-2015-large}, have been 
used for fact verification,
yielding demonstrably strong results
in the FEVER shared task.
However, there has 
been limited focus 
on the explainability of these implementations. 
Recent studies \citep{gururangan-etal-2018-annotation, mccoy-etal-2019-right} have highlighted
NLI models' tendency to 
rely on spurious cues 
for entailment classification
making it important
to provide clear explanations alongside fact verification predictions.

We design and implement a new, deterministic
NLI system based on Abstract Meaning Representation (AMR), dubbed \sysname, 
and test it on 
the FEVER and AveriTeC fact-checking datasets.
AMR is a rooted, directed, acyclic graph with nodes representing concepts and edges denoting the relations \citep{banarescu-etal-2013-abstract}. This representation captures 
semantic 
relationships among entities
that can 
be difficult to identify in 
a syntactic representation \citep{ma-etal-2023-amr}. 
We 
apply an existing AMR evaluation metric \citep{cai-knight-2013-smatch},
to map Claims (e.g., \textit{X was produced Y}) to relevant Evidence (e.g., \textit{X is a film produced by Y}). We incorporate this mapping into our \sysname system 
to yield partially explainable fact verification.

We assume 
Evidence collection has 
already been completed, as our focus is on the potential for \textit{explainability} of our fact-checking results, independent of the degree of \textit{correctness} with respect to a ground truth.  
This, in fact, is the key contribution of this paper: We demonstrate that explainability is valuable regardless of performance levels. If performance is high, explainability supports an exploration of the factors contributing to the algorithm's success. If performance is low, it serves as a diagnostic tool to understand what went wrong. 
\begin{figure*}[t]
    \centering
    \small
    \input{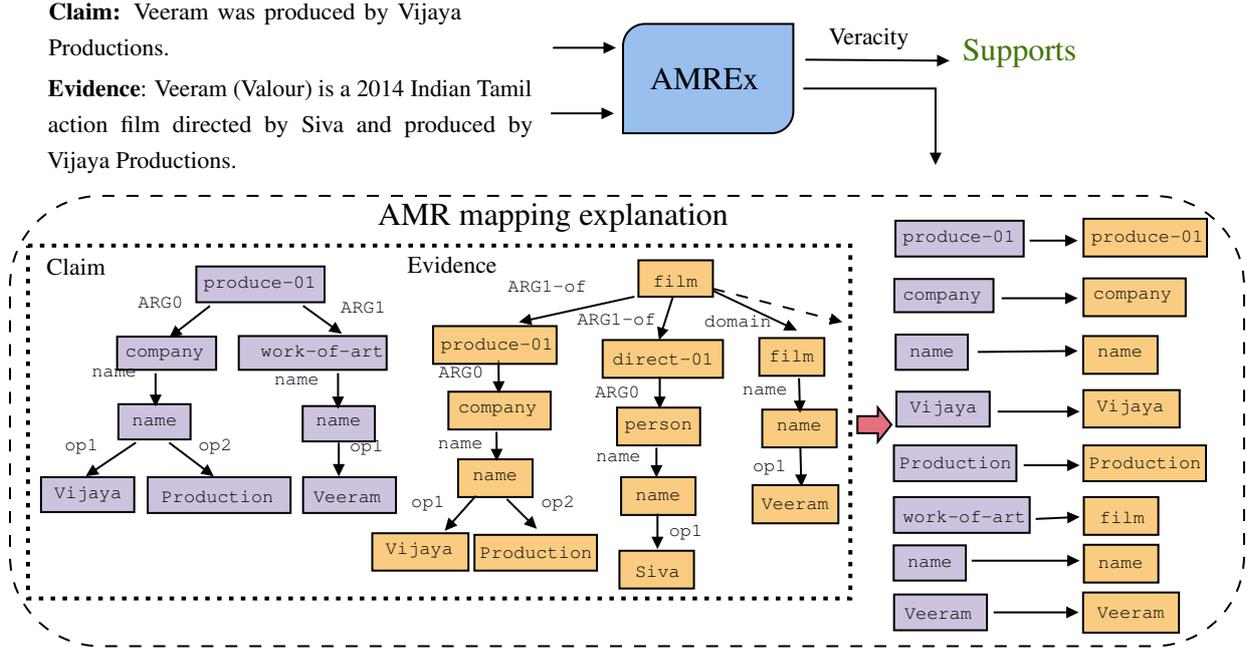}
    \caption{Explainable fact verification pipeline. Lower Left: AMR graph for the Claim, Lower Middle: AMR graph for the Evidence, Lower Right: The AMR graph mapping to explain the model's prediction as ``Supports''}
    \label{fig:explain_example}
    \vspace*{-.1in}
\end{figure*}


Fig. \ref{fig:explain_example} 
illustrates
our explainable output 
using AMR graph mapping.
We quantify the degree to which the Claim 
AMR 
is contained in the
Evidence AMR and 
present the mappings identified 
in this
process to demonstrate
whether the Claim is embedded 
within the Evidence. 
For example, the
Claim \textit{Veeram was produced by Vijay Productions} and Evidence \textit{Veeram (Valour) is a 2014 Indian Tamil action film directed by Siva and produced by Vijaya Productions} are represented as AMRs and processed through the Smatch algorithm. This 
identifies similar substructures between them,
showing 
that both texts 
mention a production (rooted by \texttt{produce-01} predicate) with similar attributes
and 
refer to
the same film 
(through substructures rooted by \texttt{work-of-art} and \texttt{film} in the Claim and Evidence AMRs). \sysname uses this high-level notion of meaning containment, 
along with a textual similarity score, to produce the veracity prediction ``Supports''. 

Section \ref{related_works}
reviews existing NLI implementations and explainable representations used in fact verification. 
Section \ref{experiment} 
provides
a detailed description of \sysname system 
and the experiments 
conducted.
Section \ref{results} 
presents an analysis and discussion of the results,
with conclusions in Section \ref{conclusion}.
    

\section{Related Work} \label{related_works}
Below 
we explore existing studies related to NLI for fact verification, Explainable representation of fact verification, and AMR. 
\subsection{NLI for Fact Verification}

NLI models have been employed for fact verification by assessing whether a given premise \textit{p} logically infers hypothesis \textit{h} \citep{bowman-etal-2015-large, zeng-zubiaga-2024-maple}. These models usually classify Claim veracity using labels: \verb|Supports|, \verb|Refutes| and \verb|NEI|. \citet{thorne-etal-2018-fact} has developed a large-scale fact verification dataset with balanced label distribution across various domains. In this study, we adopt a 3-way (FEVER) and 4-way (AVeriTec) classification for fact verification.


With the development of
fact verification datasets, fine-tuned language models (e.g., BERT, XLNet) have been applied to verify facts, 
improving generalizability without the need for manually crafted rules \citep{chernyavskiy-ilvovsky-2019-extract, nie2019revealingimportancesemanticretrieval, portelli-etal-2020-distilling, zhong-etal-2020-reasoning}. These BERT-based models use the Claim and potential Evidence as inputs 
and 
determine the final labels. Recently, \citet{pan-etal-2023-investigating} 
fine-tuned a small 
dataset to enhance the performance of BERT-based models, 
aiming to develop domain-specific models and improving generalizability. We transcend this work by employing semantic similarity 
in the embedding space between Claim and Evidence, along
with structural similarity. 

Using 
pre-trained models, graph neural networks (GNNs) have been employed to 
enhance reasoning for fact verification \citep{zhong-etal-2020-reasoning, zhou-etal-2019-gear}. These models represent Evidence as nodes within a graph, enabling information exchange between nodes, thereby improving reasoning capabilities to determine 
the final label. \citet{zhong-etal-2020-reasoning} use Semantic Role Labeling (SRL), assigning semantic roles to both Claim and Evidence sentences for 
graph construction. Building on
the concept of deeper reasoning for fact verification, we apply AMR to assess sentence similarities
through the lens of sentence structure.

Large Language Models (LLMs)
have been utilized for fact verification by augmenting verification sources. LLMs 
enable more 
realistic fact verification by considering
the date of Claims and 
using only the information available prior to the Claim \citep{chen-etal-2024-complex}. LLMs generate Claim-focused summaries, which are then used as inputs for classifiers to determine the veracity of Claims \citep{zhao-etal-2024-pacar}. Although 
LLMs have
demonstrated improved performance in fact verification, 
they still rely on classifiers that operate based on the outputs of a \textit{black box} model.

\subsection{Explainable Representations on Fact Verification}

Creating explainable justifications for fact verification predictions is an essential aspect of the task as it highlights the reasons behind a veracity prediction and presents it comprehensibly and faithfully. Several attempts have been made to create such explanations using varying techniques such as interpretable knowledge graph-based rules, attention weights, and natural-language explanations using extractive and abstractive summarization, etc. 

\citet{ahmadi2019explainable} implement an interpretable veracity prediction pipeline using Knowledge Graphs (KG) and probabilistic answer set programming that handles the uncertainties in rules created based on KGs and facts mined from the web. The resulting explanations are not in natural language but still possess a degree of interpretability. \citet{lu-li-2020-gcan}  implement a graph-based fact verification model with attention-based explanations that highlight evidential words and users when detecting fake news in tweets. Natural logic theorem proving \citep{krishna-etal-2022-proofver} produces structured explanations using an alignment-based method similar to AMREx, but it operates at the sentential level, whereas AMREx uses semantic representations to create alignments. \sysname focuses on relationships among textual entities through node mapping. Similarly, \citet{10.1145/3437963.3441828} combine structural knowledge with text embeddings to generate natural language explanations, akin to AMREx. However, their approach introduces a black-box relationship between the prediction process and explanation generation. 

Recent developments in language models have paved the way for natural-language explanation generations where both extractive and abstractive summarization are utilized for creating explanations. \citet{atanasova-etal-2020-generating-fact} train a joint model for explanation generation and veracity classification where the extractive explanations are created by selecting the most relevant ruling comments out of a collection of them for a given Claim while \citet{kotonya-toni-2020-explainable-automated} further extends this technique to create abstractive summaries for health-related Claims. Even though Large Language Models (LLMs) possess impressive generation capabilities \citet{kim2024can} show that zero-shot prompting of LLMs returns erroneous explanations due to hallucinations and focuses on generating faithful explanations using a multi-agent refinement feedback system. To address these shortcomings of LLMs, \sysname uses a linguistic approach to create a mapping of AMR graphs that explains our model's veracity predictions. We also show the potential of the mapping to be used as a prompt to generate natural-language explanations.  




\vspace{-0.21in}
\subsection{Abstract Meaning Representation (AMR)}
AMR is a rooted, directed, and acyclic semantic representation that captures the meaning of a text through
concepts and the relations that connect them \citep{banarescu-etal-2013-abstract}. It has been used for various NLP applications such as text summarization, argument similarity detection, aspect-based sentiment classification, and natural language inference \citep{dohare2017text, opitz-etal-2021-explainable, ma-etal-2023-amr, opitz-etal-2023-amr4nli}, due to its ability to capture 
key relationships among entities and generalize meaning regardless of syntax.
In \sysname, 
we focus on measuring the similarity between
two AMRs using the Smatch score \citep{cai-knight-2013-smatch}, 
which is designed to identify structural similarities of AMRs, 
effectively
comparing concept relations between pairs of texts. 

\section{Experiment} \label{experiment}
This section presents the details behind datasets used in our experiments, along with 
the experimental steps carried out to build \sysname model. 
\subsection{Datasets}
We use two fact-checking datasets to test the 
effectiveness 
of our model in verifying
the veracity of Claims, as described below.
For both datasets, we assume 
the gold Evidence for each Claim has been collected 
and thus focus only on verifying the Claim's veracity.

\subsubsection{FEVER dataset}
The FEVER dataset \citep{thorne-etal-2018-fever} consists of more than 1.8k Claims generated by altering sentences from Wikipedia. These Claims
are classified into three classes: \verb|Supports| (``S''), \verb|Refutes| (``R'') and \verb|NotEnoughInfo| (``N''). The dataset includes
relevant Evidence from Wikipedia articles for Claims 
in the first two classes. Some 
Claims require multi-hop inference/reasoning to 
verify their veracity. 

\subsubsection{AVeriTeC dataset}
AVeriTeC \citep{schlichtkrull2024averitec} is a 
newly released dataset 
containing 4568 real-world Claims. This dataset
addresses several issues associated with
previous datasets, such as 
inclusion of Evidence published after the Claim and artificially generated Claims. 
The 
Claims fall into four categories:
\verb|Supported| (``S''), \verb|Refuted| (``R''), \verb|NotEnoughEvidence| (``N'') and \verb|ConflictingEvidence/Cherrypicking| (``C''),
where 
\verb|ConflictingEvidence/Cherrypicking| represents Claims that have both supporting and refuting Evidence. 
Unlike previous datasets, 
AVeriTeC 
employs a question-answering approach to build the reasoning process for fact verification, 
encouraging
researchers to 
formulate
questions that support
Evidence extraction and to find their answers 
on the web. 

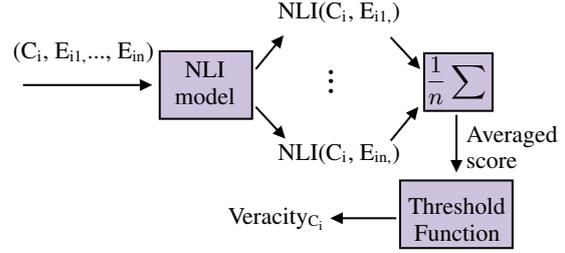
\begin{figure}[h]
    \centering
    \small
    \tikzset{every picture/.style={line width=0.75pt}} 

\begin{tikzpicture}[x=0.75pt,y=0.75pt,yscale=-1,xscale=1]

\draw  [fill={rgb, 255:red, 169; green, 150; blue, 198 }  ,fill opacity=0.57 ] (77.88,97) -- (124,97) -- (124,131) -- (77.88,131) -- cycle ;

\draw    (9.61,114) -- (71.19,114) ;
\draw [shift={(74.19,114)}, rotate = 180] [fill={rgb, 255:red, 0; green, 0; blue, 0 }  ][line width=0.08]  [draw opacity=0] (6.25,-3) -- (0,0) -- (6.25,3) -- cycle    ;
\draw    (125.85,106) -- (137.79,91.33) ;
\draw [shift={(139.69,89)}, rotate = 129.14] [fill={rgb, 255:red, 0; green, 0; blue, 0 }  ][line width=0.08]  [draw opacity=0] (6.25,-3) -- (0,0) -- (6.25,3) -- cycle    ;
\draw    (125.85,125) -- (138.57,138.79) ;
\draw [shift={(140.61,141)}, rotate = 227.31] [fill={rgb, 255:red, 0; green, 0; blue, 0 }  ][line width=0.08]  [draw opacity=0] (6.25,-3) -- (0,0) -- (6.25,3) -- cycle    ;
\draw    (193.19,90) -- (205.92,103.79) ;
\draw [shift={(207.95,106)}, rotate = 227.31] [fill={rgb, 255:red, 0; green, 0; blue, 0 }  ][line width=0.08]  [draw opacity=0] (6.25,-3) -- (0,0) -- (6.25,3) -- cycle    ;
\draw    (193.19,142) -- (205.14,127.33) ;
\draw [shift={(207.03,125)}, rotate = 129.14] [fill={rgb, 255:red, 0; green, 0; blue, 0 }  ][line width=0.08]  [draw opacity=0] (6.25,-3) -- (0,0) -- (6.25,3) -- cycle    ;
\draw  [fill={rgb, 255:red, 169; green, 150; blue, 198 }  ,fill opacity=0.57 ] (209.87,98) -- (244,98) -- (244,127) -- (209.87,127) -- cycle ;
\draw    (226,129) -- (226,155) ;
\draw [shift={(226,158)}, rotate = 270] [fill={rgb, 255:red, 0; green, 0; blue, 0 }  ][line width=0.08]  [draw opacity=0] (6.25,-3) -- (0,0) -- (6.25,3) -- cycle    ;
\draw  [fill={rgb, 255:red, 169; green, 150; blue, 198 }  ,fill opacity=0.57 ] (197.88,162) -- (254,162) -- (254,196) -- (197.88,196) -- cycle ;

\draw    (196,181) -- (166,181) ;
\draw [shift={(163,181)}, rotate = 360] [fill={rgb, 255:red, 0; green, 0; blue, 0 }  ][line width=0.08]  [draw opacity=0] (6.25,-3) -- (0,0) -- (6.25,3) -- cycle    ;

\draw (3.21,91) node [anchor=north west][inner sep=0.75pt]   [align=left] {{\fontfamily{ptm}\selectfont {\footnotesize (C\textsubscript{i}, E\textsubscript{i1,}..., E\textsubscript{in})}}};
\draw (100.48,113) node   [align=left] {\begin{minipage}[lt]{29.48pt}\setlength\topsep{0pt}
\begin{center}
{\fontfamily{ptm}\selectfont {\footnotesize NLI }}\\{\fontfamily{ptm}\selectfont {\footnotesize model}}
\end{center}

\end{minipage}};
\draw (135.44,71) node [anchor=north west][inner sep=0.75pt]   [align=left] {{\fontfamily{ptm}\selectfont {\footnotesize NLI(C\textsubscript{i}, E\textsubscript{i1,})}}};
\draw (135.44,142) node [anchor=north west][inner sep=0.75pt]   [align=left] {{\fontfamily{ptm}\selectfont {\footnotesize NLI(C\textsubscript{i}, E\textsubscript{in,})}}};
\draw (165.89,103.4) node [anchor=north west][inner sep=0.75pt]  [font=\Large,rotate=-89.31] [align=left] {...};
\draw (226.4,179) node   [align=left] {\begin{minipage}[lt]{36.18pt}\setlength\topsep{0pt}
\begin{center}
{\fontfamily{ptm}\selectfont {\footnotesize Threshold Function}}
\end{center}

\end{minipage}};
\draw (229.44,133) node [anchor=north west][inner sep=0.75pt]   [align=left] {{\fontfamily{ptm}\selectfont {\footnotesize Averaged }}\\{\fontfamily{ptm}\selectfont {\footnotesize score}}};
\draw (110.44,174) node [anchor=north west][inner sep=0.75pt]   [align=left] {{\footnotesize {\fontfamily{ptm}\selectfont Veracity}\textsubscript{C\textsubscript{i}}}};
\draw (209,98.4) node [anchor=north west][inner sep=0.75pt]    {${\displaystyle \frac{1}{n}\sum }$};

\end{tikzpicture}
    \caption{\sysname model: The model aggregates all the entailment predictions from the NLI model for a claim and returns the final veracity prediction}
    \label{fig:amrex_model}
\end{figure}
\subsection{\sysname Model}
We present the design of the \sysname 
for verification of 
Claim veracity. The underlying model is an NLI model based on a combination of an AMR evaluation metric and cosine similarity on SBERT \citep{reimers-gurevych-2019-sentence} embeddings that predicts entailment for a single (Claim, Evidence) pair. These predictions are then aggregated per claim to predict the veracity. The last stage of the model is customized to suit the different dataset formats. (See Fig. \ref{fig:amrex_model} for overall \sysname pipeline).

\subsubsection{NLI model}
\begin{figure}[h]
    \centering
    \small
    \begin{adjustwidth*}{2.5em}{2em}
    \input{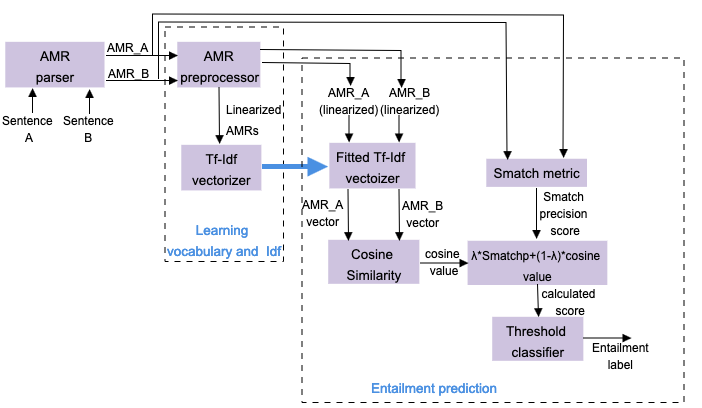}
    \end{adjustwidth*}
    \caption{NLI model pipeline. \(S_A\) refers to SentenceA and \(S_B\) refers to SentneceB. \(g_A\) refers to AMR graphA from \(S_A\) and \(g_B\) refers to AMR graphB from \(S_B\)}
    \label{fig:nli_model}
\end{figure}

Although semantic entailment does not always correspond to a strict subsumption relationship between sentences, we adopt a simplifying assumption that entailment aligns with subsumption. Specifically, our NLI model is 
based on the hypothesis that if SentenceA (\(s_A\)) semantically entails SentenceB (\(s_B\)), then the meaning of \(s_B\) 
is contained inside that of \(s_A\). This simplification allows our implementation to be built upon structured semantic concepts. Mapping this to AMR graph representations where \(g_A\) and \(g_B\) are the respective representations for \(s_A\) and \(s_B\), we hypothesize that \(g_B\) 
is a subset of \(g_A\).
To assess how much of
\(g_B\)'s meaning is contained in \(g_A\), 
we use the Smatch \citep{cai-knight-2013-smatch} precision score between \(g_A\) and \(g_B\),
combined
with the cosine similarity of SBERT embeddings of \(s_A\) and \(s_B\) (as shown in Eq. \ref{Eq:combinationeq}) to calculate the entailment score (\(f(s_A, s_B)\)) between \(s_A\) and \(s_B\). Note that the Smatch precision score is asymmetrical. So,
\(s_A\) is considered 
the premise 
and \(s_B\), the hypothesis. 
We then apply a threshold function (See Eq. \ref{Eq:threshold1}) to the resulting score to 
classify \(s_A\) as either entailing
(+1) or not entailing
(-1) 
 \(s_B\), as shown in Eq. \ref{Eq:nlieq} (See Fig. \ref{fig:nli_model}). 

\begin{equation}
\begin{split} 
\label{Eq:combinationeq}
 f(s_A, s_B) & = \lambda*Smatch_P(g_A, g_B) + \\
 & (1-\lambda)*{Cosine}_{SBERT}(s_A, s_B)
\end{split}
\end{equation}
\begin{align}
\label{Eq:threshold1}
th_1(f(s_A, s_B)) = 
 \begin{cases}
 +1, & f(s_A, s_B) \geq 0.6 \\
 -1, & f(s_A, s_B) < 0.6 \\
 \end{cases}\\
NLI(s_A, s_B) = th_1\Bigl(f(s_A, s_B)\Bigr) \label{Eq:nlieq}
\end{align}

However, as the two datasets use 
slightly different labeling schemes
(FEVER uses 
a 3-way classification format, while AVeriTeC 
uses a 4-way classification format) and a Claim may involve 
multiple pieces of Evidence 
in the entailment process, the fact verification 
approach needs to be customized for each dataset. 
This customization will be described 
in Sections \ref{fever_verification} and \ref{averitec_verification}. As observed in this implementation, minor variations in verdict labels may exist across different datasets, we believe these differences are not substantial, as all labels pertain to assessing the truth value of a claim. Therefore, the threshold function can be readily adjusted to accommodate new verdict labels.



\begin{table}[] 

\resizebox{\columnwidth}{!}{%
\begin{tabular}{llllll}
\hline
\textbf{Dataset} & \textbf{S} & \textbf{R} & \textbf{N} & \begin{tabular}[c]{@{}l@{}}\textbf{C}\end{tabular} & \begin{tabular}[c]{@{}l@{}}\textbf{Total} \\ \textbf{\# sentences}\end{tabular} \\ \hline
FEVER & 3281 & 3270 & 3284 & - & 9835 \\
AVeriTec & 649 & 1166 & 115 & 226 & 2156 \\ \hline
\end{tabular}%
}
\caption{Label distribution of FEVER and AVeriTec datasets: Supports (S), Refutes (R), Not Enough Evidence (N), Conflicting Evidence (C).}
\label{table:labeldist}
\end{table}

\subsubsection{Fact verification for FEVER} \label{fever_verification}
The FEVER dataset categorizes 
Claims and Evidence 
into three classes (\verb|Supports|, \verb|Refutes|, \verb|NotEnoughInfo|. Each Claim may have 
one or more
pieces of Evidence, while those labeled NEI 
lack any Evidence. 
To address the lack 
of Evidence for the NEI class, we use the modified FEVER dataset provided by \citet{atanasova-etal-2020-generating}, which includes
Evidence for NEI class. 

\begin{table*}[h]
\centering
\begin{tabular}{llllllll}
\hline
\textbf{Model}                                                         & \textbf{lambda} & \textbf{S} & \textbf{R} & \textbf{N} & \textbf{C} & \textbf{\begin{tabular}[c]{@{}l@{}}Macro\\ F1\end{tabular}} & \textbf{Acc.} \\ \hline
\textbf{FEVER baseline}                                                           & \_              & \_         & \_         & \_         & \_         & \_                                                          & 0.88          \\
\begin{tabular}[c]{@{}l@{}}\(AMREx_{FEVER_{acc, f1}}\)\end{tabular}    & 0               & 0.52           & 0.39           & 0.41            & \_           & 0.44                                                            & 0.44              \\
\textbf{AVeriTec baseline}                                                        & \_              & 0.48       & 0.74       & 0.59       & 0.15       & 0.49                                                        & 0.49          \\
\begin{tabular}[c]{@{}l@{}}\(AMREx_{AVeriTec_{acc}}\)\end{tabular} & 0.9             & 0.10       & 0.67       & 0.04       & 0.02       & 0.21                                                        & \textbf{0.50} \\
\begin{tabular}[c]{@{}l@{}}\(AMREx_{AVeriTec_{f1}}\)\end{tabular}   & 0               & 0.25       & 0.61       & 0.06       & 0.11       & 0.26                                                        & 0.43          \\ \hline
\end{tabular}
\caption{Accuracy and Macro F1 scores of veracity prediction for each veracity label. Only accuracy is reported in FEVER baseline.}
\label{table:result}
\end{table*}

Given a pair (\(C_i\), \(E_{ij}\)) where \(C_i\) is a Claim and \(E_{ij}\) is its jth Evidence,  we use the NLI pipeline shown in Fig. \ref{fig:nli_model} to compute the entailment between them. Here, 
\(C_i\) is treated as the hypothesis and  \(E_{ij}\) as the premise. If \(E_{ij}\) entails \(C_i\), 
it returns +1. If not,
it returns -1, as outlined
in Eq. \ref{Eq:nlieq}. \sysname then
averages the results across
all Evidence 
for \(C_i\) from the NLI model, 
to determine the overall entailment (e), and classify that into one of the three classes 
using a threshold classifier to return the veracity of \(C_i\), as shown in Eq. \ref{Eq:th2} and \ref{Eq:veracity}. When deciding the thresholds for the labels, “Supports” and “Refutes” are given the positive and negative extremes, respectively, whereas “Not enough Info” is assigned the middle range. This is based on the assumption that evidence with insufficient information will exhibit lower structural and textual similarity scores without extreme contradictions. The exact threshold values were determined experimentally.

\begin{align}
 th_{2_{FV}}(e) = 
 \begin{cases}
 \text{``S''}, & e \geq 0.1 \\
 \text{``N''}, & -0.1 < e < 0.1 \\
 \text{``R''}, & e \leq -0.1
 \end{cases} \label{Eq:th2}
\vspace*{-.25in}
\end{align}

\begin{align}    
\vspace*{-.25in}
\label{Eq:veracity}
 Veracity_{C_{i}} = th_{2_{FV}}\Bigl(\frac{1}{n}\sum_{j=1}^{n}NLI(C_i, E_{ij})\Bigr)
\end{align}

\vspace{-0.15in}
\subsubsection{Fact verification for AVeriTeC} 
\label{averitec_verification}
The AVeriTeC dataset requires 
a Claim extraction system to first create questions to aid in finding Evidence related to a Claim, and then locate 
relevant documents and sentences 
to answer those questions, which are considered 
Evidence for the Claim. Since
we assume 
the correct questions and answers are already provided
for each Claim, we calculate the overall entailment between a Claim and Evidence using Eq. \ref{Eq:veracity}. However,
we apply a customized threshold function for the AVeriTec dataset 
as it includes 
four veracity labels (Eq. \ref{Eq:threshold_averitec}). 
Additionally, the dataset 
features three types of Evidence: Boolean, Abstractive, and Extractive.
Since 
Boolean Evidence (Yes/No answers) is
incompatible with both AMRs and our entailment pipeline, 
we focus on 
abstractive and extractive Evidence in the experiment to fully measure our pipeline's ability to represent
sentential Evidence. 
Table \ref{table:labeldist} shows the label distribution of both datasets.

\begin{align}    
 th_{2_{AV}}(e) = 
 \begin{cases}
 \text{``S''}, & e \geq 0.5 \\
 \text{``C''}, & 0.1 < e < 0.5\\
 \text{``N''}, & -0.1 \leq e \leq 0.1 \\
 \text{``C''}, & -0.5 < e < -0.1\\
 \text{``R''}, & e\leq -0.5
 \end{cases}
 \label{Eq:threshold_averitec}
\end{align}

\section{Results and Analyses} \label{results}
We experiment
with \(\lambda\) values in the [0,1] range for Eq. \ref{Eq:combinationeq} 
on both FEVER and AVeriTec datasets to 
find the best combination of AMR graph intersection 
and textual similarity measurement. The results for both datasets are in Table \ref{table:result}. We selected the best-performing models 
based on both the highest accuracy and macro F1 score, leading to two
\sysname implementations 
for each dataset.

For
the FEVER dataset, 
the best
accuracy and macro F1 score are achieved when \(\lambda = 0\), suggesting that the Smatch precision score 
has minimal impact on
predicting the veracity 
of (Claim, Evidence) pairs.
The label-wise performance shows that \(AMREx_{FEVER{acc,f1}}\) 
is more effective at
identifying supporting (Claim, Evidence) pairs but struggles with 
refuting instances.

However, the 
AVeriTec dataset exhibits
different behavior, with
\(\lambda = 0.9\) yielding the best accuracy
and \(\lambda = 0\) producing 
the best macro F1 score. \(AMREx_{AVeriTec_{acc}}\) also manages to surpass the AVeriTec accuracy baseline. \(AMREx_{AVeriTec_{f1}}\) performs
comparably 
to the AVeriTec baseline in recognizing refutable (Claim, Evidence) pairs and those with conflicting evidence. However, with greater emphasis on
the Smatch precision score when \(\lambda = 0.9\), \(AMREx_{AVeriTec_{f1}}\) 
improves
in identifying refutable (Claim, Evidence) pairs, albeit 
at the 
cost of performance on other label instances. 

Through an error analysis, we identify several cases where \sysname fails to accurately predict the veracity 
and we explore
their potential causes. Consider the following supporting (Claim, Evidence) pair from the FEVER dataset, Claim: \textit{``Wish Upon was released in the 21st century.''}, Evidence: \textit{``It is set to be released in theaters on July 14, 2017, by Broad Green Pictures and Orion Pictures''} (See Fig. \ref{fig:amr-rep-erroranalysis-1} for corresponding AMRs in Penman notation \citep{goodman-2020-penman}). \sysname returns the following mapping for this instance with a Smatch precision score of 0.53 and a textual similarity score of 0.38.

\noindent\texttt{a0(release-01) --> b2(release-01) \\
a1(music) --> b1(it)\\ 
a2(name) --> b10(name)\\
a3(Wish) --> b11(Orion)\\
a4(Upon) --> b12(Pictures)\\
a5(date-entity) --> b14(date-entity)}

\begin{figure}[h]
    \centering
\vspace*{-0.1in}
\begin{lstlisting}[basicstyle=\footnotesize\ttfamily,xleftmargin=0em,framexleftmargin=4em,firstnumber=1] 
AMR Corresponding to the Claim:
(a0/release-01
   :ARG1 (a1/music
      :name (a2/name
         :op1 (a3/Wish)
         :op2 (a4/Upon)))
   :time (a5/date-entity
      :century 21))

AMR Corresponding to the Evidence:
(b0/set-08
   :ARG1 (b1/it)
   :ARG2 (b2/release-01
      :ARG0 (b3/and
         :op1 (b4/company
            :name (b5/name
               :op1 (b6/Broad)
               :op2 (b7/Green)
               :op3 (b8/Pictures)))
         :op2 (b9/company
            :name (b10/name
               :op1 (b11/Orion)
               :op2 (b12/Pictures))))
      :ARG1 i
      :location (b13/theater)
      :time (b14/date-entity
         :day 14
         :month 7
         :year 2017)))
\end{lstlisting}
\caption{Abstract Meaning Representations (AMRs) for Claim: ``Wish Upon was released in the 21st century.'' and Evidence: ``It is set to be released in theaters on July 14, 2017, by Broad Green Pictures and Orion Pictures''}
\label{fig:amr-rep-erroranalysis-1}
\end{figure}

The AMR node mapping correctly identifies that both texts are related to a release event (with the \texttt{a0} node mapping to the \texttt{b2} node), connects ``music'' in Claim AMR 
to ``it'' in Evidence AMR, and 
recognizes that both texts 
mention a \texttt{date-entity}. 
However, it 
fails to map ``the 21st-century'' 
in the Claim
with the 
date in the Evidence AMR. The Smatch precision score 
indicates a higher level of
meaning entailment compared to the textual similarity score,
but it is not 
high enough to 
meet the entailment threshold with any \(\lambda\) value,
leading 
\sysname to incorrectly predict ``Refutes''. This 
reveals a limitation of the Smatch algorithm 
in inferring 
that the year 2017 
falls within the 21st century, as
it is a concept mapping algorithm. 
We note that SBERT contextual embeddings also fail to capture this detail and give an even 
lower similarity assessment.

Another 
example reveals that 
high structural similarity between AMRs, 
despite a few factual differences, 
can 
result in incorrect meaning containment assessments. Consider the Claim: \textit{``Marnie is a romantic film.''} and the Evidence: \textit{``Marnie is a 1964 American psychological thriller film directed by Alfred Hitchcock.''} with the gold veracity label ``Refutes'' (See Fig. \ref{fig:amr-rep-erroranalysis-2} for AMRs). The resulting AMR node mappings are as follows:

\noindent\texttt{a0(film) --> b0(film)\\ a1(romantic-03) --> b1(direct-01)\\ a2(name) --> b11(name)\\ a3(Marnie) --> b12(Marnie)}
\begin{figure}[h]
    \centering
\begin{lstlisting}[basicstyle=\footnotesize\ttfamily,xleftmargin=0em,framexleftmargin=4em,firstnumber=1] 
AMR Corresponding to the Claim:
(a0/film
   :ARG0-of (a1/romantic-03)
   :name (a2/name
      :op1 (a3/Marnie)))

AMR Corresponding to the Evidence:
(b0/film
   :ARG1-of (b1/direct-01
      :ARG0 (b2/person
         :name (b3/name
            :op1 (b4/Alfred)
            :op2 (b5/Hitchcock))))
   :mod (b6/thriller
      :mod (b7/psychological))
   :mod (b8/country
      :name (b9/name
         :op1 (b10/America)))
   :name (b11/name
      :op1 (b12/Marnie))
   :time (b13/date-entity
      :year 1964))
\end{lstlisting}
\caption{Abstract Meaning Representations (AMRs) for Claim: ``Marnie is a romantic film.'' and Evidence: ``Marnie is a 1964 American psychological thriller film directed by Alfred Hitchcock.''}
\label{fig:amr-rep-erroranalysis-2}
\end{figure}

In the Claim AMR, ``Marnie'' being a ``romantic film'' is represented by the 
\texttt{romantic-03} node, 
while in the Evidence, it being a ``Psychological thriller'' is represented 
by a modifier to the root \texttt{film}. Due to this structural discrepancy,
the Smatch algorithm fails to
distinguish between the two genres and instead maps 
\texttt{romantic-03} to \texttt{direct-01} with a similar structure that still correctly creates 
a mismatch, but for the wrong reason. 
However, most concepts in 
the Claim AMR 
match those 
in the Evidence AMR,
leading to 
a high Smatch precision score of 0.75. The textual similarity score also returns a 0.70. Hence,
any \(\lambda\) combination of the two scores surpasses the entailment threshold, yielding a 
``Supports'' prediction. 

These examples reveal that the 
AMR and textual similarity-based approach
of \sysname struggles
with instances 
involving implied meaning or those with high structural similarity 
but factual differences, indicating areas that need improvement.

\subsection{Explainability of the Model}
The model's explainability 
stems from two key aspects.  
First, the deterministic nature of the model's calculations 
allows us to trace 
how a particular prediction was calculated. 
This provides a comprehensive explanation of the entire system pipeline and tracks the process at each step. 
Second, 
the 
visual mapping between the AMRs of Claims and Evidence, as shown in Fig. \ref{fig:explain_example}, helps 
clarify why the
model returns a particular prediction 
for a (Claim, Evidence) pair in terms of structural similarity. This explanation is partial and post hoc, relying only on AMR node mappings for generation. However, it is integrated into the system, as AMR representations influence both the veracity prediction and explanation generation. An example illustrating AMREx’s explanations is discussed below.
\begin{figure}[h]
    \centering
\begin{lstlisting}[basicstyle=\footnotesize\ttfamily,xleftmargin=0em,framexleftmargin=4em,firstnumber=1] 
AMR Corresponding to the Claim:
(a0/ride-01
   :ARG1 (a1/disease
      :name (a2/name
         :op1 (a3/Rabies)))
   :location (a4/amusement-park
      :name (a5/name
         :op1 6
         :op2 (a6/Parks))))

AMR Corresponding to the Evidence:
(b0/disease
   :ARG0-of (b1/cause-01
      :ARG1 (b2/inflame-01
         :ARG1 (b3/brain)
         :part-of (b4/and
            :op1 (b5/human)
            :op2 (b6/mammal
               :mod (b7/other)))))
   :domain (b8/disease
      :name (b9/name
         :op1 (b10/Rabies)))
   :mod (b11/virus))
\end{lstlisting}
\caption{Abstract Meaning Representations (AMRs) for Claim: ``Rabies is a ride at Six Parks.'' and Evidence: ``Rabies is a viral disease that causes inflammation of the brain in humans and other mammals.''}
\label{fig:amr-rep}
\vspace*{-.15in}
\end{figure}

\begin{figure}[t]
    \centering
    \scriptsize
    \tikzset{every picture/.style={line width=0.75pt}} 

\begin{tikzpicture}[x=0.76pt,y=0.63pt,yscale=-1,xscale=1]

\draw   (3.5,0.63) -- (291.16,0.63) -- (291.16,659) -- (3.5,659) -- cycle ;

\draw (147.34,329.81) node   [align=left] {\begin{minipage}[lt]{196.38pt}\setlength\topsep{0pt}
{\fontfamily{ptm}\selectfont {\footnotesize \textbf{AMR Graph Mapping Analysis:}}}\\...\\{\fontfamily{ptm}\selectfont {\footnotesize \textbf{Key Mappings:}}}\\\\
{\footnotesize {\fontfamily{pcr}\selectfont a0(ride-01)-b0(disease)}{\fontfamily{ptm}\selectfont : The concept "ride" in the claim is mapped to "disease" in the evidence, which indicates a mismatch. The claim is about a ride, whereas the evidence is about a disease.}}\\\\{\footnotesize {\fontfamily{pcr}\selectfont a1(disease)-b8(disease): }{\fontfamily{ptm}\selectfont Both the claim and the evidence refer to "Rabies" as a disease. However, in the claim, it's tied to a "ride" rather than a health context.}}\\\\{\footnotesize {\fontfamily{pcr}\selectfont a2(name)-b9(name): }{\fontfamily{ptm}\selectfont The name "Rabies" is consistently mapped in both the claim and the evidence.}}\\\\{\footnotesize {\fontfamily{pcr}\selectfont a3(Rabies)-b10(Rabies): }{\fontfamily{ptm}\selectfont The specific instance "Rabies" is correctly identified in both the claim and the evidence.}}\\\\{\footnotesize {\fontfamily{pcr}\selectfont a4(amusement-park)-b2(inflame-01):}{\fontfamily{ptm}\selectfont  The location "amusement park" in the claim is mapped to "inflammation" in the evidence, which again indicates a significant mismatch in context.}}\\\\{\footnotesize {\fontfamily{pcr}\selectfont a5(name)-b4(and),a6(Parks)-b6(mammal): }{\fontfamily{ptm}\selectfont The reference to "Six Parks" in the claim is incorrectly mapped to a reference to "mammals" in the evidence.}}\\\\{\fontfamily{ptm}\selectfont {\footnotesize \textbf{Explanation:}}}\\\\
{\fontfamily{ptm}\selectfont {\footnotesize \textbf{Mismatch in Core Concepts:} The claim describes "Rabies" as a "ride" at an "amusement park," whereas the evidence discusses "Rabies" as a "viral disease." The AMR mappings reveal that the central concepts of "ride" and "amusement park" in the claim do not align with the "disease" and "medical context" in the evidence.}}\\ ... \\
{\fontfamily{ptm}\selectfont {\footnotesize \textbf{Classification:} Given that the evidence contradicts the central concept of the claim by presenting a different context (medical vs. amusement), this pair should be classified as REFUTES. The evidence does not support the claim that "Rabies is a ride at Six Parks" and instead presents a fact that contradicts this claim.}}

\end{minipage}};
\vspace*{-.2in}
\end{tikzpicture}
    \vspace*{-.2in}
    \caption{Natural Language explanation generated by ChatGPT based on the \sysname's AMR node mapping output.}
    \label{fig:generated_explanation}
    \vspace*{-.2in}
\end{figure}

Consider Claim \textit{``Rabies is a ride at Six Parks.''} and the Evidence, \textit{``Rabies is a viral disease that causes inflammation of the brain in humans and other mammals.''} The corresponding AMRs for Claim and Evidence are shown in Fig. \ref{fig:amr-rep}. 
When these two AMRs are processed through the Smatch algorithm, the resulting AMR node mapping is 
as follows:

\noindent\texttt{a0(ride-01) --> b0(disease) \\
a1(disease) --> b8(disease) \\
a2(name) --> b9(name) \\
a3(Rabies) --> b10(Rabies) \\
a4(amusement-park) --> b2(inflame-01) \\
a5(name) --> b4(and) \\
a6(Parks) --> b6(mammal)}

As the mapping reveals, the only shared meaning 
between these two AMRs is that both sentences are related to a disease 
called Rabies (with the \texttt{a1-a3} nodes mapping to the \texttt{b8-b10} nodes). This leads to 
a low Smatch precision score of 0.46 and a textual similarity score of 0.59. When 
combined with any \(\lambda\) value, this results in a low entailment value, causing the threshold function to
predict non-entailment (-1). Hence, the second threshold function will return ``Refutes'' as the veracity of this (Claim, Evidence) pair. 

This 
process, together with the AMR node mapping, demonstrates the model's overall logic 
and final prediction. We further demonstrate the effectiveness of the AMR node mapping 
in generating natural-language explanations by using it 
to prompt ChatGPT to generate justifications. An excerpt of the generated explanation for the (Claim, Evidence) pair above is shown in Fig. \ref{fig:generated_explanation}. Hence, 
we argue that 
\sysname's explainable output can 
serve as a guide for generating natural-language explanations for veracity detection, helping to reduce hallucinations 
in LLM models.

\section{Conclusion} \label{conclusion}
We implement \sysname, an Abstract Meaning Representation-based veracity prediction and explanation system for fact verification, and show its effectiveness in producing explainable justifications using two fact verification datasets. 
Although its performance is lower compared to the baselines, its partially explainable output could still be used as a diagnostic tool to perform error analyses on the veracity prediction system to understand the areas to improve. We further demonstrate that \sysname output can 
guide
LLMs to generate natural-language explanations using the AMR mappings.


\section*{Limitations}
In its current form, \sysname performs best across all classes when its score is less influenced by the structural similarity assessment. This makes the AMR node mappings less useful as explanations, despite achieving the highest performance. Therefore, further exploration is needed to adjust the structural similarity assessment to better suit the veracity detection task. Since \sysname relies heavily on AMRs,
it is crucial to use a 
high-performing AMR parser when converting the sentences to AMRs. Therefore, the overall performance of the system depends on the accuracy of the AMR parser. Furthermore, the AMR mapping algorithm is 
more effective when applied to
text instances 
with a high degree of structural similarity,
which may not always be the case 
with real-world data. The AMR node mappings provide a partial, post hoc explanation of the system, while the interpretability of the entire system fully encompasses the prediction process. 
An evaluation of the explainable aspect of AMREx model 
in comparison to current structural explainable fact verification systems is also necessary. We expect to address these limitations in future modifications to the system.

\section*{Ethical Statement}
We utilize ChatGPT responses as a demonstration of the effectiveness of \sysname in creating
natural-language explanations for veracity predictions. We acknowledge that there is a possibility for ChatGPT to generate hallucinated, or toxic content. However, one of the key objectives of our study is to develop an explainable system 
whose output can 
guide the reduction of hallucinations 
in LLM-generated outputs, including ChatGPT. We believe this approach contributes to the generation of content that is both faithful and safe. 
Additionally, we manually check the ChatGPT-generated content in this study for 
hallucinated or toxic content and can 
confirm that the presented examples are 
free of such issues.

\section*{Acknowledgement}
This material is based upon work supported, in part, by the Defense Advanced Research Projects Agency (DARPA) under Contract No. HR001121C0186. Any opinions, findings and conclusions or recommendations expressed in this material are those of the authors and do not necessarily reflect the views of the Defense Advanced Research Projects Agency (DARPA).
\bibliography{anthology,custom}

\begin{thebibliography}{34}
\expandafter\ifx\csname natexlab\endcsname\relax\def\natexlab#1{#1}\fi

\bibitem[{Ahmadi et~al.(2019)Ahmadi, Lee, Papotti, and Saeed}]{ahmadi2019explainable}
Naser Ahmadi, Joohyung Lee, Paolo Papotti, and Mohammed Saeed. 2019.
\newblock Explainable fact checking with probabilistic answer set programming.
\newblock In \emph{Conference on Truth and Trust Online}.

\bibitem[{Atanasova et~al.(2020{\natexlab{a}})Atanasova, Simonsen, Lioma, and Augenstein}]{atanasova-etal-2020-generating-fact}
Pepa Atanasova, Jakob~Grue Simonsen, Christina Lioma, and Isabelle Augenstein. 2020{\natexlab{a}}.
\newblock \href {https://doi.org/10.18653/v1/2020.acl-main.656} {Generating fact checking explanations}.
\newblock In \emph{Proceedings of the 58th Annual Meeting of the Association for Computational Linguistics}, pages 7352--7364. Association for Computational Linguistics.

\bibitem[{Atanasova et~al.(2020{\natexlab{b}})Atanasova, Wright, and Augenstein}]{atanasova-etal-2020-generating}
Pepa Atanasova, Dustin Wright, and Isabelle Augenstein. 2020{\natexlab{b}}.
\newblock \href {https://doi.org/10.18653/v1/2020.emnlp-main.256} {Generating label cohesive and well-formed adversarial claims}.
\newblock In \emph{Proceedings of the 2020 Conference on Empirical Methods in Natural Language Processing (EMNLP)}, pages 3168--3177. Association for Computational Linguistics.

\bibitem[{Banarescu et~al.(2013)Banarescu, Bonial, Cai, Georgescu, Griffitt, Hermjakob, Knight, Koehn, Palmer, and Schneider}]{banarescu-etal-2013-abstract}
Laura Banarescu, Claire Bonial, Shu Cai, Madalina Georgescu, Kira Griffitt, Ulf Hermjakob, Kevin Knight, Philipp Koehn, Martha Palmer, and Nathan Schneider. 2013.
\newblock \href {https://aclanthology.org/W13-2322} {{A}bstract {M}eaning {R}epresentation for sembanking}.
\newblock In \emph{Proceedings of the 7th Linguistic Annotation Workshop and Interoperability with Discourse}, pages 178--186. Association for Computational Linguistics.

\bibitem[{Bowman et~al.(2015)Bowman, Angeli, Potts, and Manning}]{bowman-etal-2015-large}
Samuel~R. Bowman, Gabor Angeli, Christopher Potts, and Christopher~D. Manning. 2015.
\newblock \href {https://doi.org/10.18653/v1/D15-1075} {A large annotated corpus for learning natural language inference}.
\newblock In \emph{Proceedings of the 2015 Conference on Empirical Methods in Natural Language Processing}, pages 632--642. Association for Computational Linguistics.

\bibitem[{Cai and Knight(2013)}]{cai-knight-2013-smatch}
Shu Cai and Kevin Knight. 2013.
\newblock \href {https://aclanthology.org/P13-2131} {{S}match: an evaluation metric for semantic feature structures}.
\newblock In \emph{Proceedings of the 51st Annual Meeting of the Association for Computational Linguistics (Volume 2: Short Papers)}, pages 748--752. Association for Computational Linguistics.

\bibitem[{Chen et~al.(2024)Chen, Kim, Sriram, Durrett, and Choi}]{chen-etal-2024-complex}
Jifan Chen, Grace Kim, Aniruddh Sriram, Greg Durrett, and Eunsol Choi. 2024.
\newblock \href {https://doi.org/10.18653/v1/2024.naacl-long.196} {Complex claim verification with evidence retrieved in the wild}.
\newblock In \emph{Proceedings of the 2024 Conference of the North American Chapter of the Association for Computational Linguistics: Human Language Technologies (Volume 1: Long Papers)}, pages 3569--3587. Association for Computational Linguistics.

\bibitem[{Chernyavskiy and Ilvovsky(2019)}]{chernyavskiy-ilvovsky-2019-extract}
Anton Chernyavskiy and Dmitry Ilvovsky. 2019.
\newblock \href {https://doi.org/10.18653/v1/D19-6612} {Extract and aggregate: A novel domain-independent approach to factual data verification}.
\newblock In \emph{Proceedings of the Second Workshop on Fact Extraction and VERification (FEVER)}, pages 69--78. Association for Computational Linguistics.

\bibitem[{Dohare et~al.(2017)Dohare, Karnick, and Gupta}]{dohare2017text}
Shibhansh Dohare, Harish Karnick, and Vivek Gupta. 2017.
\newblock Text summarization using abstract meaning representation.
\newblock \emph{arXiv preprint arXiv:1706.01678}.

\bibitem[{Goodman(2020)}]{goodman-2020-penman}
Michael~Wayne Goodman. 2020.
\newblock \href {https://doi.org/10.18653/v1/2020.acl-demos.35} {{P}enman: An open-source library and tool for {AMR} graphs}.
\newblock In \emph{Proceedings of the 58th Annual Meeting of the Association for Computational Linguistics: System Demonstrations}, pages 312--319. Association for Computational Linguistics.

\bibitem[{Gururangan et~al.(2018)Gururangan, Swayamdipta, Levy, Schwartz, Bowman, and Smith}]{gururangan-etal-2018-annotation}
Suchin Gururangan, Swabha Swayamdipta, Omer Levy, Roy Schwartz, Samuel Bowman, and Noah~A. Smith. 2018.
\newblock \href {https://doi.org/10.18653/v1/N18-2017} {Annotation artifacts in natural language inference data}.
\newblock In \emph{Proceedings of the 2018 Conference of the North {A}merican Chapter of the Association for Computational Linguistics: Human Language Technologies, Volume 2 (Short Papers)}, pages 107--112. Association for Computational Linguistics.

\bibitem[{Kim et~al.(2024)Kim, Lee, Huang, Chan, Li, and Ji}]{kim2024can}
Kyungha Kim, Sangyun Lee, Kung-Hsiang Huang, Hou~Pong Chan, Manling Li, and Heng Ji. 2024.
\newblock Can llms produce faithful explanations for fact-checking? towards faithful explainable fact-checking via multi-agent debate.
\newblock \emph{arXiv preprint arXiv:2402.07401}.

\bibitem[{Kotonya and Toni(2020)}]{kotonya-toni-2020-explainable-automated}
Neema Kotonya and Francesca Toni. 2020.
\newblock \href {https://doi.org/10.18653/v1/2020.emnlp-main.623} {Explainable automated fact-checking for public health claims}.
\newblock In \emph{Proceedings of the 2020 Conference on Empirical Methods in Natural Language Processing (EMNLP)}, pages 7740--7754. Association for Computational Linguistics.

\bibitem[{Krishna et~al.(2022)Krishna, Riedel, and Vlachos}]{krishna-etal-2022-proofver}
Amrith Krishna, Sebastian Riedel, and Andreas Vlachos. 2022.
\newblock \href {https://doi.org/10.1162/tacl_a_00503} {{P}roo{FV}er: Natural logic theorem proving for fact verification}.
\newblock \emph{Transactions of the Association for Computational Linguistics}, 10:1013--1030.

\bibitem[{Lu and Li(2020)}]{lu-li-2020-gcan}
Yi-Ju Lu and Cheng-Te Li. 2020.
\newblock \href {https://doi.org/10.18653/v1/2020.acl-main.48} {{GCAN}: Graph-aware co-attention networks for explainable fake news detection on social media}.
\newblock In \emph{Proceedings of the 58th Annual Meeting of the Association for Computational Linguistics}, pages 505--514. Association for Computational Linguistics.

\bibitem[{Ma et~al.(2023)Ma, Hu, Liu, Yang, Li, Yu, and Wen}]{ma-etal-2023-amr}
Fukun Ma, Xuming Hu, Aiwei Liu, Yawen Yang, Shuang Li, Philip~S. Yu, and Lijie Wen. 2023.
\newblock \href {https://doi.org/10.18653/v1/2023.acl-long.19} {{AMR}-based network for aspect-based sentiment analysis}.
\newblock In \emph{Proceedings of the 61st Annual Meeting of the Association for Computational Linguistics (Volume 1: Long Papers)}, pages 322--337. Association for Computational Linguistics.

\bibitem[{McCoy et~al.(2019)McCoy, Pavlick, and Linzen}]{mccoy-etal-2019-right}
Tom McCoy, Ellie Pavlick, and Tal Linzen. 2019.
\newblock \href {https://doi.org/10.18653/v1/P19-1334} {Right for the wrong reasons: Diagnosing syntactic heuristics in natural language inference}.
\newblock In \emph{Proceedings of the 57th Annual Meeting of the Association for Computational Linguistics}, pages 3428--3448. Association for Computational Linguistics.

\bibitem[{Nie et~al.(2019)Nie, Wang, and Bansal}]{nie2019revealingimportancesemanticretrieval}
Yixin Nie, Songhe Wang, and Mohit Bansal. 2019.
\newblock \href {http://arxiv.org/abs/1909.08041} {Revealing the importance of semantic retrieval for machine reading at scale}.
\newblock In \emph{Proceedings of the 2019 Conference on Empirical Methods in Natural Language Processing}. Association for Computational Linguistics.

\bibitem[{Opitz et~al.(2021)Opitz, Heinisch, Wiesenbach, Cimiano, and Frank}]{opitz-etal-2021-explainable}
Juri Opitz, Philipp Heinisch, Philipp Wiesenbach, Philipp Cimiano, and Anette Frank. 2021.
\newblock \href {https://doi.org/10.18653/v1/2021.argmining-1.3} {Explainable unsupervised argument similarity rating with {A}bstract {M}eaning {R}epresentation and conclusion generation}.
\newblock In \emph{Proceedings of the 8th Workshop on Argument Mining}, pages 24--35. Association for Computational Linguistics.

\bibitem[{Opitz et~al.(2023)Opitz, Wein, Steen, Frank, and Schneider}]{opitz-etal-2023-amr4nli}
Juri Opitz, Shira Wein, Julius Steen, Anette Frank, and Nathan Schneider. 2023.
\newblock \href {https://aclanthology.org/2023.iwcs-1.29} {{AMR}4{NLI}: Interpretable and robust {NLI} measures from semantic graphs}.
\newblock In \emph{Proceedings of the 15th International Conference on Computational Semantics}, pages 275--283. Association for Computational Linguistics.

\bibitem[{Pan et~al.(2023)Pan, Zhang, and Kan}]{pan-etal-2023-investigating}
Liangming Pan, Yunxiang Zhang, and Min-Yen Kan. 2023.
\newblock \href {https://doi.org/10.18653/v1/2023.ijcnlp-main.34} {Investigating zero- and few-shot generalization in fact verification}.
\newblock In \emph{Proceedings of the 13th International Joint Conference on Natural Language Processing and the 3rd Conference of the Asia-Pacific Chapter of the Association for Computational Linguistics (Volume 1: Long Papers)}, pages 511--524. Association for Computational Linguistics.

\bibitem[{Portelli et~al.(2020)Portelli, Zhao, Schuster, Serra, and Santus}]{portelli-etal-2020-distilling}
Beatrice Portelli, Jason Zhao, Tal Schuster, Giuseppe Serra, and Enrico Santus. 2020.
\newblock \href {https://doi.org/10.18653/v1/2020.fever-1.7} {Distilling the evidence to augment fact verification models}.
\newblock In \emph{Proceedings of the Third Workshop on Fact Extraction and VERification (FEVER)}, pages 47--51. Association for Computational Linguistics.

\bibitem[{Reimers and Gurevych(2019)}]{reimers-gurevych-2019-sentence}
Nils Reimers and Iryna Gurevych. 2019.
\newblock \href {https://doi.org/10.18653/v1/D19-1410} {Sentence-{BERT}: Sentence embeddings using {S}iamese {BERT}-networks}.
\newblock In \emph{Proceedings of the 2019 Conference on Empirical Methods in Natural Language Processing and the 9th International Joint Conference on Natural Language Processing (EMNLP-IJCNLP)}, pages 3982--3992. Association for Computational Linguistics.

\bibitem[{Schlichtkrull et~al.(2024)Schlichtkrull, Guo, and Vlachos}]{schlichtkrull2024averitec}
Michael Schlichtkrull, Zhijiang Guo, and Andreas Vlachos. 2024.
\newblock Averitec: A dataset for real-world claim verification with evidence from the web.
\newblock \emph{Advances in Neural Information Processing Systems}, 36.

\bibitem[{Schlichtkrull et~al.(2023)Schlichtkrull, Guo, and Vlachos}]{schlichtkrull2023averitec}
Michael~Sejr Schlichtkrull, Zhijiang Guo, and Andreas Vlachos. 2023.
\newblock \href {https://openreview.net/forum?id=fKzSz0oyaI} {Averitec: A dataset for real-world claim verification with evidence from the web}.
\newblock In \emph{Thirty-seventh Conference on Neural Information Processing Systems Datasets and Benchmarks Track}.

\bibitem[{Thorne et~al.(2018{\natexlab{a}})Thorne, Vlachos, Christodoulopoulos, and Mittal}]{thorne-etal-2018-fever}
James Thorne, Andreas Vlachos, Christos Christodoulopoulos, and Arpit Mittal. 2018{\natexlab{a}}.
\newblock \href {https://doi.org/10.18653/v1/N18-1074} {{FEVER}: a large-scale dataset for fact extraction and {VER}ification}.
\newblock In \emph{Proceedings of the 2018 Conference of the North {A}merican Chapter of the Association for Computational Linguistics: Human Language Technologies, Volume 1 (Long Papers)}, pages 809--819. Association for Computational Linguistics.

\bibitem[{Thorne et~al.(2018{\natexlab{b}})Thorne, Vlachos, Cocarascu, Christodoulopoulos, and Mittal}]{thorne-etal-2018-fact}
James Thorne, Andreas Vlachos, Oana Cocarascu, Christos Christodoulopoulos, and Arpit Mittal. 2018{\natexlab{b}}.
\newblock \href {https://doi.org/10.18653/v1/W18-5501} {The fact extraction and {VER}ification ({FEVER}) shared task}.
\newblock In \emph{Proceedings of the First Workshop on Fact Extraction and {VER}ification ({FEVER})}, pages 1--9. Association for Computational Linguistics.

\bibitem[{Thorne et~al.(2019)Thorne, Vlachos, Cocarascu, Christodoulopoulos, and Mittal}]{thorne-etal-2019-fever2}
James Thorne, Andreas Vlachos, Oana Cocarascu, Christos Christodoulopoulos, and Arpit Mittal. 2019.
\newblock \href {https://doi.org/10.18653/v1/D19-6601} {The {FEVER}2.0 shared task}.
\newblock In \emph{Proceedings of the Second Workshop on Fact Extraction and VERification (FEVER)}, pages 1--6. Association for Computational Linguistics.

\bibitem[{Vedula and Parthasarathy(2021)}]{10.1145/3437963.3441828}
Nikhita Vedula and Srinivasan Parthasarathy. 2021.
\newblock \href {https://doi.org/10.1145/3437963.3441828} {Face-keg: Fact checking explained using knowledge graphs}.
\newblock In \emph{Proceedings of the 14th ACM International Conference on Web Search and Data Mining}, WSDM '21, page 526–534, New York, NY, USA. Association for Computing Machinery.

\bibitem[{Vlachos and Riedel(2014)}]{vlachos-riedel-2014-fact}
Andreas Vlachos and Sebastian Riedel. 2014.
\newblock \href {https://doi.org/10.3115/v1/W14-2508} {Fact checking: Task definition and dataset construction}.
\newblock In \emph{Proceedings of the {ACL} 2014 Workshop on Language Technologies and Computational Social Science}, pages 18--22. Association for Computational Linguistics.

\bibitem[{Zeng and Zubiaga(2024)}]{zeng-zubiaga-2024-maple}
Xia Zeng and Arkaitz Zubiaga. 2024.
\newblock \href {https://aclanthology.org/2024.findings-eacl.79} {{MAPLE}: Micro analysis of pairwise language evolution for few-shot claim verification}.
\newblock In \emph{Findings of the Association for Computational Linguistics: EACL 2024}, pages 1177--1196. Association for Computational Linguistics.

\bibitem[{Zhao et~al.(2024)Zhao, Wang, Wang, Cheng, Zhang, and Wong}]{zhao-etal-2024-pacar}
Xiaoyan Zhao, Lingzhi Wang, Zhanghao Wang, Hong Cheng, Rui Zhang, and Kam-Fai Wong. 2024.
\newblock \href {https://aclanthology.org/2024.lrec-main.1099} {{PACAR}: Automated fact-checking with planning and customized action reasoning using large language models}.
\newblock In \emph{Proceedings of the 2024 Joint International Conference on Computational Linguistics, Language Resources and Evaluation (LREC-COLING 2024)}, pages 12564--12573. ELRA and ICCL.

\bibitem[{Zhong et~al.(2020)Zhong, Xu, Tang, Xu, Duan, Zhou, Wang, and Yin}]{zhong-etal-2020-reasoning}
Wanjun Zhong, Jingjing Xu, Duyu Tang, Zenan Xu, Nan Duan, Ming Zhou, Jiahai Wang, and Jian Yin. 2020.
\newblock \href {https://doi.org/10.18653/v1/2020.acl-main.549} {Reasoning over semantic-level graph for fact checking}.
\newblock In \emph{Proceedings of the 58th Annual Meeting of the Association for Computational Linguistics}, pages 6170--6180. Association for Computational Linguistics.

\bibitem[{Zhou et~al.(2019)Zhou, Han, Yang, Liu, Wang, Li, and Sun}]{zhou-etal-2019-gear}
Jie Zhou, Xu~Han, Cheng Yang, Zhiyuan Liu, Lifeng Wang, Changcheng Li, and Maosong Sun. 2019.
\newblock \href {https://doi.org/10.18653/v1/P19-1085} {{GEAR}: Graph-based evidence aggregating and reasoning for fact verification}.
\newblock In \emph{Proceedings of the 57th Annual Meeting of the Association for Computational Linguistics}, pages 892--901. Association for Computational Linguistics.

\end{thebibliography}
\bibliographystyle{acl_natbib}

\end{document}